\documentclass[]{TEAI}

\usepackage{helvet}

\usepackage{amsmath} 
\usepackage{mathtools}
\usepackage{natbib}
\usepackage{graphicx}
\usepackage{subcaption} 

\usepackage[toc,page,header]{appendix}
\usepackage[utf8]{inputenc} 
\usepackage[T1]{fontenc}    
\usepackage{hyperref}       
\usepackage{url}            
\usepackage{booktabs}       
\usepackage{lmodern}        
\usepackage{amsfonts}       
\usepackage{nicefrac}       
\usepackage{microtype}      
\usepackage{wrapfig}

\usepackage{amssymb}  
\usepackage{fontawesome}  
\usepackage{url}  

\usepackage{titletoc}

\usepackage{tikz}  
\usepackage{comment}  
\usepackage{tabularx}  
\usepackage{booktabs}  

\usepackage{minitoc}

\usepackage{booktabs}
\usepackage{array}
\usepackage{etoolbox}

\definecolor{lightblue}{RGB}{200, 230, 255}  
\definecolor{headerblue}{RGB}{150, 200, 255} 

\usepackage{pgfplots}
\usepackage[utf8]{inputenc} 
\usepackage[T1]{fontenc}    
\usepackage{hyperref}       
\usepackage{url}            
\usepackage{booktabs}       
\usepackage{amsfonts}       
\usepackage{nicefrac}       
\usepackage{microtype}      
\usepackage{graphicx}
\usepackage{float}
\usepackage{comment}
\usepackage{multirow} 
\usepackage{amsmath} 
\usepackage{makecell} 
\usepackage{siunitx}  
\usepackage{tikz}
\usepackage{pgf-pie} 
\usepackage{subcaption}
\usepackage{wrapfig}
\usepackage[export]{adjustbox}

\usepackage{ragged2e}      
\usepackage{tabularx}       
\usepackage{array}          
\usepackage{caption}        
\usepackage{enumitem}
\usepackage{pifont}
\usepackage[hang,flushmargin]{footmisc} 

\usepackage[ruled,vlined]{algorithm2e}

\usepackage{tcolorbox}

\usepackage{tcolorbox}    
\tcbuselibrary{breakable}  
\tcbuselibrary{skins}      

\usepackage{tabularx}
\usepackage{listings}

\definecolor{mygreen}{RGB}{76, 200, 80}  
\definecolor{mygray}{RGB}{240, 240, 240} 
\definecolor{myred}{RGB}{244, 67, 54}     

\newcommand{\OurFramework}{\texttt{DeepForge}}
\newcommand{\OurDataset}{\texttt{DeepForge-Dataset}}

\newcommand{\OurQADataset}{\texttt{DeepForge-QA}}

\newcommand{\OurModel}{\texttt{OffSeeker}}

\tcbset{
  prompt/.style={
    width=\linewidth,
    top=8pt,
    bottom=4pt,
    colback=SeaGreen!10!CornflowerBlue!10,
    colframe=black,
    colbacktitle=black,
    enhanced,
    center,
    attach boxed title to top left={yshift=-0.1in,xshift=0.15in},
    boxed title style={boxrule=0pt,colframe=white,},
  }
}

\newtcolorbox{PromptBox}[2][]{prompt,breakable,title=#2,#1}

\newtcolorbox{turnbox}[1][]{
  colback=blue!5!white, colframe=blue!75!black, boxrule=0.5pt, arc=3pt, left=2pt, right=2pt, top=2pt, bottom=2pt, enhanced, width=0.25\textwidth,
  #1
}

\title{\textsc{OffSeeker}: Online Reinforcement Learning Is Not All You Need for Deep Research Agents}

\author{
    Yuhang Zhou\textsuperscript{1,2},
    Kai Zheng\textsuperscript{2,*},
    Qiguang Chen\textsuperscript{1},  
    Mengkang Hu\textsuperscript{3},  \\
    Qingfeng Sun\textsuperscript{2},
    Can Xu\textsuperscript{2,$\dagger$}
    Jingjing Chen\textsuperscript{1,$\dagger$}
}

\affiliation[1]{\mbox{Fudan University}} 
\affiliation[2]{\mbox{Tencent Hunyuan}}
\affiliation[3]{\mbox{The University of Hong Kong}}

\correspondence{\textcolor{seedblue}{ralph.yh.zhou@gmail.com, \{kevinezheng,leocaxu\}@tencent.com, chenjingjing@fudan.edu.cn}}
\checkdata[Website]{\url{https://github.com/Ralph-Zhou/OffSeeker}}

\begin{document}
\abstract{
\begin{abstract}

Deep research agents have shown remarkable potential in handling long-horizon tasks. 
However, state-of-the-art performance typically relies on online reinforcement learning (RL), which is financially expensive due to extensive API calls. While offline training offers a more efficient alternative, its progress is hindered by the scarcity of high-quality research trajectories. In this paper, we demonstrate that \textit{expensive online reinforcement learning is not all you need} to build powerful research agents. To bridge this gap, we introduce a fully open-source suite designed for effective offline training. Our core contributions include \OurFramework{}, a ready-to-use task synthesis framework that generates large-scale research queries without heavy preprocessing; and a curated collection of 66k QA pairs, 33k SFT trajectories, and 21k DPO pairs. Leveraging these resources, we train \OurModel{} (8B), a model developed entirely offline.
Extensive evaluations across six benchmarks show that \OurModel{} not only leads among similar-sized agents but also remains competitive with 30B-parameter systems trained via heavy online RL.




\end{abstract}
}

\maketitle
\renewcommand{\thefootnote}{}
\footnotetext{$^*$Project Lead.\\$^\dagger$Corresponding authors.}
\renewcommand{\thefootnote}{\arabic{footnote}}


\section{Introduction}
Large language model (LLM)-based agents have been widely applied to complex tasks involving extended reasoning, multi-step decision-making, and tool use~\citep{gao2025survey, hu2025owl, team2025kimi, fang2025cognitive, chen2025towards, hu2025agentgen}.
Among them, deep research agents specialize in long-horizon web information gathering, iterative hypothesis refinement, and structured synthesis, and have achieved strong performance on challenging benchmarks such as GAIA, BrowseComp, Xbench, and HLE~\citep{mialon2023gaia, wei2025browsecomp, chen2025xbench, phan2025humanity}.
Despite this progress, developing deep research agents remains difficult in practice due to (1) inefficient open-source data synthesis frameworks, (2) limited availability of high-quality training trajectories, and (3) the high cost of online reinforcement learning, which together hinder scalability, accessibility, and reproducibility in academic research.

Specifically, as summarized in Table~\ref{tab:opensource_comparison}, most systems employ online RL with live web interactions. This causes instability in  non-stationary environments and incurs high API costs (over \$350 for 50 GRPO steps), undermining research priorities such as cost control, reproducibility, and accessibility, and confining large-scale experiments to well-resourced industry labs.

In contrast, offline training is constrained by inadequate data synthesis frameworks and datasets. A key bottleneck lies in obtaining diverse seed entities, as most methods depend on well-cleaned and well-structured Wikipedia resources~\citep{liu2025webexplorer,gao2025beyond,li2025websailor,lu2025deepdive}, which demand heavy human effort for preprocessing and yield limited timeliness and diversity.
Constrained by these frameworks, high-quality trajectories and preference data remain scarce. Existing deep-research datasets lack the complexity and diversity required for advanced reasoning~\citep{hu2025owl}.

\begin{table*}[t!]
    \centering
    \caption{Comparison among recent deep research works. ``Training Regime'' indicates whether model training follows an online (``On'') or offline (``Off'') procedure. ``API Cost'' denotes the estimated monetary cost (USD) arising from additional search API calls during the complete training pipeline due to interactions with external environments. More details about our estimation method can be found in Appendix \ref{appendix:api-cost}.}
    \scalebox{0.8}{
    \begin{tabular}{
        m{4cm} 
        >{\centering\arraybackslash}m{2.6cm}
        >{\centering\arraybackslash}m{1.7cm}
        >{\centering\arraybackslash}m{1.4cm}
        >{\centering\arraybackslash}m{2.2cm} 
        >{\centering\arraybackslash}m{2.3cm}
        >{\centering\arraybackslash}m{1.6cm}
        >{\centering\arraybackslash}m{1.4cm}
    }
    \toprule
    \multirow{3}{*}{\textbf{Current Works}} & \multicolumn{5}{c}{\textbf{Open Source Components}} & \multirow{3}{*}{\makecell[c]{\textbf{Training}\\\textbf{Regime}}} & \multirow{3}{*}{\makecell[c]{\textbf{API}\\\textbf{Cost}}} \\
    \cmidrule(lr){2-6}
    & \textbf{Data Synthesis Framework} & \textbf{Training Script}  & \textbf{Model Weight} & \textbf{Deep Search QA Data} & \textbf{High-Quality Trajectories} & & \\
    \midrule
    Kimi-K2~\citep{team2025kimi} & \textcolor{myred}{\ding{55}} & \textcolor{myred}{\ding{55}} & \textcolor{mygreen}{\checkmark} & - & - & Off+On & - \\
    WebDancer~\citep{wu2025webdancer} & \textcolor{myred}{\ding{55}} & \textcolor{myred}{\ding{55}} & \textcolor{mygreen}{\checkmark} & 200 & 200 & Off+On & - \\
    WebSailor~\citep{li2025websailor} & \textcolor{myred}{\ding{55}} & \textcolor{myred}{\ding{55}} & \textcolor{mygreen}{\checkmark} & 20 & - & Off+On & - \\
    WebShaper~\citep{tao2025webshaper} & \textcolor{myred}{\ding{55}} & \textcolor{myred}{\ding{55}} & \textcolor{mygreen}{\checkmark} & 500 & - & Off & 0 \\
    WebExplorer~\citep{liu2025webexplorer} & \textcolor{myred}{\ding{55}} & \textcolor{myred}{\ding{55}} & \textcolor{mygreen}{\checkmark} & 100 & - & Off+On & $>$\$2,918 \\
    Tongyi DeepResearch~\citep{team2025tongyi} & \textcolor{myred}{\ding{55}} & \textcolor{myred}{\ding{55}} & \textcolor{mygreen}{\checkmark} & - & - & Off+On & - \\
    DeepDive~\citep{lu2025deepdive} & \textcolor{mygreen}{\checkmark} & \textcolor{myred}{\ding{55}} & \textcolor{myred}{\ding{55}} & 3.2k & 858 & Off+On & $>$\$4,300 \\
    Asearcher-v1~\citep{gao2025beyond} & \textcolor{mygreen}{\checkmark} & \textcolor{mygreen}{\checkmark} & \textcolor{mygreen}{\checkmark} & 35k & - & On & $>$\$360 \\
    \rowcolor{blue!5}{\textbf{Ours}} & \textcolor{mygreen}{\checkmark} & \textcolor{mygreen}{\checkmark} & \textcolor{mygreen}{\checkmark} & \textbf{66k} & \textbf{33k} & \textbf{Off} & \textbf{0} \\
    \bottomrule
    \end{tabular}
    }
    \label{tab:opensource_comparison}
    \vspace{-0.3cm}
\end{table*}

To solve this, our work systematically addresses the three key challenges described above by providing a comprehensive, fully open-source deep research agent training solution. 

First, we introduce \OurFramework{}, an end-to-end pipeline for synthesizing deep research tasks through real-world web interaction.
It uses LLM-based prompting and iterative expansion to construct large collections of initial entities and URLs, together with entity graph.
This lightweight design eliminates the need for heavy preprocessing and enables the synthesis of dynamic and diverse tasks beyond static resources such as Wikipedia.

Second, to address the scarcity of academic resources, we release the first comprehensive large-scale deep research dataset to the public, including 66k challenging QA pairs generated by \OurFramework{}, 33k SFT trajectories, and 21k DPO preference pairs. This dataset substantially lowers the entry barrier for research in this area.

Third, we demonstrate that \textit{\textbf{online reinforcement learning is not all you need}} to develop enterprise-level offline-trained deep research agents. We present \OurModel{}, an 8B-parameter deep research agent based on Qwen3-8B~\citep{yang2025qwen3}, trained entirely using offline supervised fine-tuning and DPO. Experimental results show that our offline approach matches or even exceeds the performance of prominent online RL systems, for example, 26.6 on BrowseComp-zh, compared with 25.5 for WebSailor-32B, while significantly reducing training costs and providing better reproducibility.


Our contributions are as follows:
\begin{itemize}[leftmargin=16pt,itemsep=0pt,topsep=0pt]
    \item We open-source \OurFramework{}, an end-to-end framework for synthesizing large-scale and diverse deep research tasks that require extensive multi-step reasoning and exploration.
    The framework is ready to use out of the box and does not require heavy preprocessing.    
    \item We release the first fully open deep-research resource suite, including the full codebase and dataset: 66k challenging deep research tasks, 33k high-quality SFT trajectories, and 21k DPO preference data, substantially lowering barriers to academic research.
    \item We release our offline-trained 8B model, \OurModel{}, via a fully offline training recipe. It achieves strong deep-research capabilities without online RL, reducing API costs to near zero. Experiments show \OurModel{} ranks top-tier among comparable-scale agents and rivals 30B systems trained with online RL.
\end{itemize}

\section{\texttt{DeepForge}: Large Scale Deep-Research Data Synthesis}
\label{sec:data-generation}

\begin{figure*}[h]
    \centering
    \includegraphics[width=\textwidth]{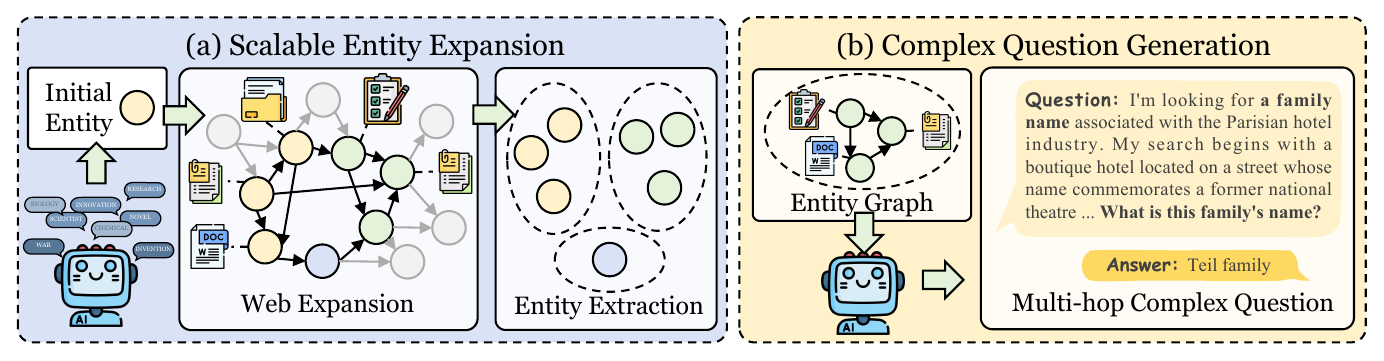}
    \vspace{-0.2cm}
    \caption{Overview of our \OurFramework{} data synthesis pipeline. \OurFramework{} comprises two main stages: (a) Scalable Entity Expansion and (b) Complex Question Generation. After synthesizing complex deep search tasks, we further deploy an agent framework to collect high-quality trajectories.}
    \label{fig:deepforge}
\end{figure*}
Existing open-source data synthesis frameworks require extensive preprocessing, such as downloading full Wikipedia dumps or querying the Wikipedia SPARQL API~\citep{liu2025webexplorer,gao2025beyond,lu2025deepdive,tao2025webshaper,li2025websailor}. These heavy requirements increase implementation complexity and constrain dataset diversity. Moreover, frameworks from ASearcher and DeepDive rely primarily on fully offline Wikipedia datasets, which cover only a small fraction of real-world web sources. Consequently, these approaches cannot generate dynamic, up-to-date high-quality data.

To address these limitations, we introduce \OurFramework{}, a simple yet highly effective pipeline that synthesizes large-scale deep research tasks and is ready to use immediately. We then construct \OurQADataset, a 66k challenging deep search dataset for model training.

\subsection{Scalable Entity Expansion}
In contrast to existing approaches that rely on extensive preprocessing infrastructures, \OurFramework{} initiates the process with a lightweight and scalable entity expansion stage. As illustrated in Figure~\ref{fig:deepforge}, we first prompt an LLM to generate a broad pool of initial entities, typically simple nouns that collectively span a wide range of domains. We then iteratively refine this pool into a comprehensive seed entity set through a two-step process: (1) For each initial entity, we query a web search API to collect a set of relevant URLs, thereby constructing a large-scale seed URL corpus; (2) For each seed URL, we extract clean textual content using \texttt{html2text}, and prompt an LLM to identify additional entities mentioned within the retrieved text. To ensure that the generated questions encourage genuine research, the LLM is instructed to prioritize the extraction of long-tail and low-frequency entities. Following deduplication and noise filtering, this procedure results in a large, diverse, and high-quality entity pool suitable for constructing multi-hop knowledge graphs.

\subsection{Complex Question Generation}
Given the expanded entity set, \OurFramework{} constructs an entity-centric knowledge graph through multi-step traversal. Starting from each seed entity, a lightweight search agent retrieves both intrinsic attributes and relational links to other entities. Recursively expanding along low-frequency or sparsely documented relations yields an entity graph with rich cross-entity dependencies.

Using this graph as context, we prompt DeepSeek-v3.1~\citep{liu2024deepseek} to synthesize questions that require multiple rounds of web search, cross-entity reasoning, and ambiguity resolution. Each question is paired with a gold answer grounded in the retrieved evidence. To further enhance difficulty, we introduce a pruning stage that removes overly specific clues while preserving solvability, intentionally reducing surface-level cues and increasing the depth of reasoning required. Using this pipeline, we construct the query of \OurQADataset, a dataset containing 66k high-difficulty, multi-hop, and high-uncertainty deep search questions. More details can be found in Appendix~\ref{appendix:data_generation_pipeline}.

\begin{table*}[t]
    \centering
    \caption{Comparison of current open-source deep search datasets statistics. Statistics from other works are obtained from~\citet{liu2025webexplorer}. For ASearcher, we report their LRM split results. For DeepDive, we calculate the average turns based on their open-source SFT trajectories.}
    \scalebox{0.76}{
    \begin{tabular}{
        m{0.32\textwidth}
        >{\centering\arraybackslash}m{0.12\textwidth}
        >{\centering\arraybackslash}m{0.1\textwidth}
        >{\centering\arraybackslash}m{0.12\textwidth}
        >{\centering\arraybackslash}m{0.11\textwidth}
        >{\centering\arraybackslash}m{0.12\textwidth}
        >{\centering\arraybackslash}m{0.12\textwidth}
        >{\centering\arraybackslash}m{0.09\textwidth}
        }
    \toprule
    \textbf{Dataset} & \textbf{WebDancer} & \textbf{SailorFog} & \textbf{WebShaper} & \textbf{ASearcher} & \textbf{DeepDive} & \textbf{WebExplorer} & \textbf{Ours} \\
    \midrule
    Number of open-source samples & 200 & 20 & 500 & 35k & 3.2k & 100 & \textbf{66k} \\
    Average Turns & 5.4 & 8.2 & 8.4 & 6.5 & 9.9 & 10.6 & \textbf{11.2} \\
    \bottomrule
    \end{tabular}
    }
    \label{tab:opensource_dataset_comparison}
\end{table*}

\subsection{Difficulty Analysis of Synthesized Tasks}
High-quality deep research tasks require extensive multi-step reasoning and exploration. To assess the quality and difficulty of tasks generated by \OurFramework{}, we randomly sample 5,000 synthesized queries. As all questions are solvable by our research agent, DeepSeek-v3.1, we compare the distribution of tool call counts as a difficulty metric~\citep{chen2024unlocking,qu2025tool}.

\begin{figure}[h]
    \begin{minipage}{0.5\columnwidth}
    As illustrated in Table~\ref{tab:opensource_dataset_comparison} and Figure~\ref{fig:tool_call_distribution}, the difficulty of \OurQADataset{} substantially exceeds that of most existing open-source deep research datasets, indicating greater suitability for training models on complex tasks. Specifically, the generated tasks demand sustained multi-step exploration, with an average of 11.2 tool calls per question, far surpassing comparable datasets such as ASearcher. These results demonstrate that \OurFramework{} produces tasks that better capture the uncertainty and long-horizon reasoning of real-world deep research problems, emphasizing long-term planning, information synthesis, and iterative evidence gathering. Case study about the generated tasks can be found in Appendix~\ref{appendix:deepforge_case_study}.
    \end{minipage}
    \hfill  
    \begin{minipage}{0.45\columnwidth}
    \centering
    \includegraphics[width=\linewidth]{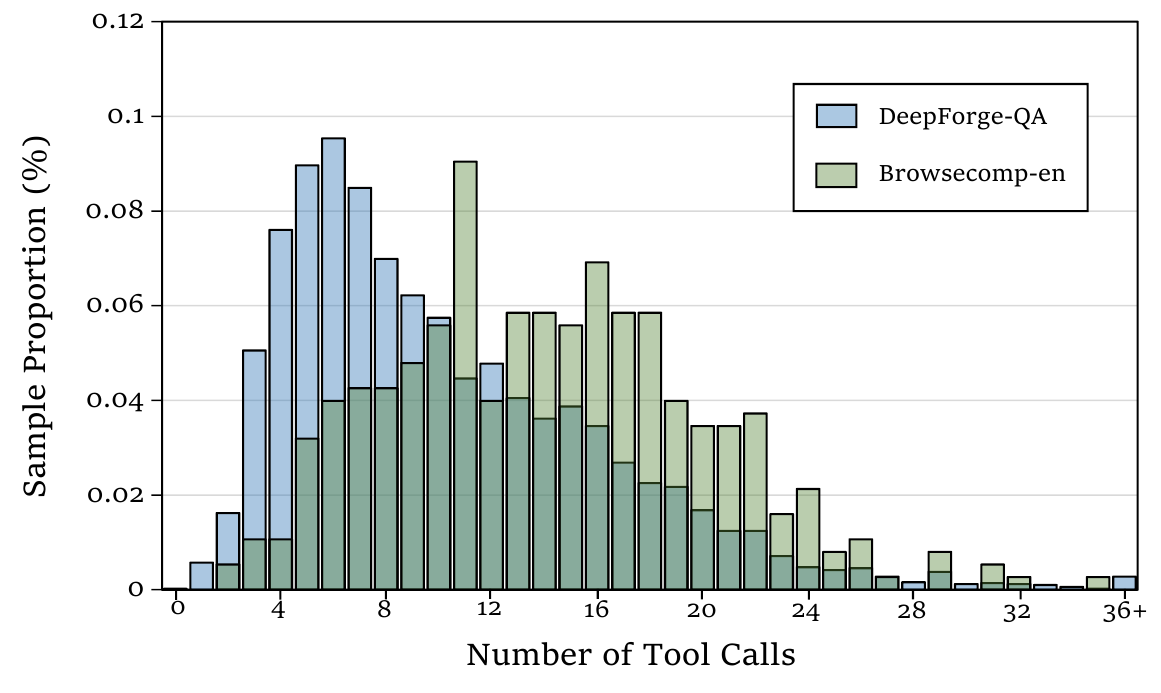}
    \caption{Comparison of tool calling turns distribution between \OurFramework{}-generated tasks and BrowseComp-en with DeepSeek-v3.1.}
    \label{fig:tool_call_distribution}
    \end{minipage}
    \hfill
\end{figure}



\section{Methodology}

\subsection{Overall Agent Framework}

We adopt a ReAct-style agent framework~\citep{yao2022react}, which explicitly interleaves reasoning and action. The agent's intermediate reasoning, tool invocations, and tool outputs are represented using \texttt{<think>}, \texttt{<tool\_call>}, and \texttt{<tool\_response>} tags, respectively. Tool responses are appended to the user context and fed back to the agent as part of the context, enabling iterative interaction with the environment.

Formally, given an initial task query $q$, we denote the agent's internal reasoning at step $t$ as $\tau_t$, the action (tool invocation) as $\alpha_t$, and the environment feedback as $o_t$. The agent iteratively reasons and interacts until producing a final answer enclosed by \texttt{<answer>}. A complete interaction trajectory is:
\[
T = \left( q, \tau_0, \alpha_0, o_0, \tau_1, \alpha_1, o_1, \ldots, \tau_t, \langle \texttt{ans} \rangle \right).
\]
To support efficient deep-research behavior, we equip the agent with four tools: \texttt{search}, \texttt{visit\_urls}, \texttt{search\_wiki}, and \texttt{code\_execution}. 
More details about the agent framework can be found in Appendix~\ref{appendix:agent-framework}.


\subsection{Supervised Fine-tuning}

\paragraph{SFT Trajectory Construction.}
To obtain high-quality step-by-step reasoning trajectories, we built an agent framework with corresponding tools and used proprietary models to sample and obtain trajectories. For each question, the agent interacts with the real Internet environment, performs multi-step exploration, retrieves evidence, and produces a final answer.
Since our constructed data primarily focuses on web search tasks, we include a small portion of open-source datasets containing other task types to mitigate catastrophic forgetting and ensure the model maintains the ability to utilize tools such as code execution.

To ensure the quality, the sampled trajectories are then filtered through a multi-stage quality pipeline that checks for: (1) invalid tool results and API errors; (2) format consistency and structural validity; (3) answer correctness through semantic equivalence evaluation; and (4) overall trajectory quality using LLM-based assessment. After filtering, we obtain \textbf{33k high-quality trajectories} for SFT training. Detailed filtering procedures can be found in Appendix~\ref{appendix:data-filtering}.

\paragraph{Training Objective.}
We perform standard supervised fine-tuning (SFT) on the curated trajectories. Formally, let $\left( x, T \right) \sim \mathcal{D}_{\mathrm{SFT}}$ denote a training example, where $x$ is the initial task query and 
$T = \left( q, \tau_0, \alpha_0, o_0, \ldots, \tau_t, \langle \texttt{ans} \rangle \right)$ 
is the full interaction trajectory containing intermediate reasoning, tool invocations, environment observations, and the final answer. We adopt the standard autoregressive SFT objective, minimizing the negative log-likelihood of the ground-truth trajectory:

\vspace{-0.3cm}
\begin{equation}
\mathcal{L}_{\mathrm{SFT}}(\theta)
= \mathbb{E}_{\left( x, T \right) \sim \mathcal{D}_{\mathrm{SFT}}}
\left[
- \sum_{t=1}^{|T|}
\log p_{\theta}(T_t \mid x, T_{<t})
\right]
\end{equation}

\subsection{Direct Preference Optimization}

To address high-cost challenges in online RL, we explore an offline alternative that avoids online interaction altogether. Specifically, we adopt Direct Preference Optimization (DPO)~\citep{rafailov2023direct}, a widely used offline RL method that operates purely on preference data.

\paragraph{Preference Data Construction.}


We first deploy the SFT-trained model to generate multiple candidate trajectories for each task. Subsequently, a powerful LLM-based evaluator is utilized to assess and assign scores to each trajectory across several dimensions, such as logical consistency, factual correctness, and overall solution quality.

For each task, we select the top two highest-scoring trajectories and the bottom two lowest-scoring trajectories, and pair them exhaustively to form four preference pairs. To ensure high-quality supervision, we further filter the data by retaining only pairs in which the preferred trajectory receives a strictly higher evaluation score than the rejected one. Following this procedure, we construct approximately 21k high-quality preference pairs for DPO training.

\paragraph{Training Objective.}
DPO encourages the model to assign higher likelihood to preferred trajectories while remaining close to the reference SFT policy. Let $(T^{+}, T^{-}) \sim \mathcal{D}_{\mathrm{DPO}}$ denote a preference pair. We define the log-ratio
\begin{equation}
\Delta_{\theta}(T) = \log \pi_{\theta}(T) - \log \pi_{\mathrm{ref}}(T),
\end{equation}
where $\pi_{\theta}$ is the current policy and $\pi_{\mathrm{ref}}$ denotes the frozen SFT reference model. The DPO objective is then given by
\begin{equation}
\mathcal{L}_{\mathrm{DPO}}(\theta)
=
- \mathbb{E}_{(T^{+}, T^{-})}
\left[
\log \sigma \big(
\beta ( \Delta_{\theta}(T^{+}) - \Delta_{\theta}(T^{-}) )
\big)
\right],
\end{equation}
where $\beta$ is the hyperparameter that controls the strength of preference optimization.

\section{Preliminary: Online RL Cost Analysis}
To assess API overhead and time costs in online RL for training deep research models, we experimented with a SFT-warmup 8B model further refined via online RL.

\paragraph{Experiment Setup.}

We used the GRPO~\citep{shao2024deepseekmath} algorithm, implemented in the rllm~\citep{rllm2025} framework, on a subset of \OurQADataset.
We tracked (1) model performance on a sampled BrowseComp-ZH subset, (2) cumulative search API costs during training, and (3) per-step training time.
More details can be found in Appendix~\ref{appendix:grpo-training-efficiency}.

\paragraph{Online RL is highly API-costly.}

\begin{figure*}[ht]
    \centering
    \includegraphics[width=0.98\textwidth]{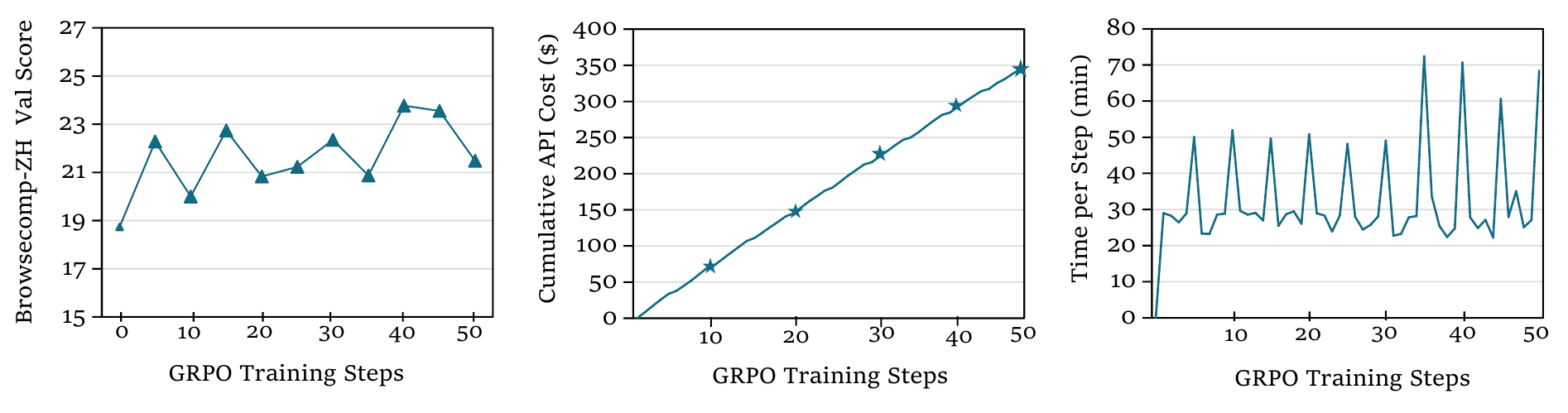}
    \vspace{-0.3cm}
    \caption{Analysis of GRPO training results. Left: Model performance (score) on BrowseComp-ZH during GRPO training. Middle: Cumulative API expenditure (in USD) throughout training. Right: Time consumption (minutes) of each training step.}
    \label{fig:grpo-experiment}
    \vspace{-0.3cm}
\end{figure*}

As shown in Figure~\ref{fig:grpo-experiment} (middle), GRPO training for 50 steps incurred $\approx\$350$ in API costs.
Tool calls and environment interactions per task also slowed rollouts substantially; a single step took about 20--30 minutes (Figure~\ref{fig:grpo-experiment}, right).
Online RL further introduces instability, as large-scale GRPO sampling causes API call bursts that risk rate limits or failures from external services.
Offline training, by contrast, incurs no API costs, enabling strong deep-research models without search API expenses.

\section{Main Experiments}

\begin{table*}[t]
    \centering
    \caption{Main results on deep research benchmarks. All scores of \OurModel{} are computed as Pass@1 using LLM-as-Judge. BC-zh and BC-en denote the BrowseComp-ZH and BrowseComp benchmarks, respectively, while XBench-DS refers to XBench-DeepSearch. Scores marked with $^\dagger$ are reported from our own tests, others are reported from official reports or other deep research works. For GAIA and HLE, we use text-only subset of the original benchmarks. \textbf{Bold} represents the best performance among \textit{Small Size Deep Research Agents} on each benchmark, while \underline{underlined} values represent the second best performance.}
    \begin{spacing}{1.05}
    \scalebox{0.95}{
    \begin{tabular}{
        l
        >{\centering\arraybackslash}m{1.5cm}
        >{\centering\arraybackslash}m{1.5cm}
        >{\centering\arraybackslash}m{1.5cm}    
        >{\centering\arraybackslash}m{1.3cm}
        >{\centering\arraybackslash}m{2.1cm}
        >{\centering\arraybackslash}m{2.3cm}
    }
    \toprule
    \textbf{Backbone}
    & \textbf{GAIA}
    & \textbf{BC-zh}
    & \textbf{BC-en}
    & \textbf{HLE}
    & \textbf{XBench-DS} 
    & \textbf{WebWalkerQA}\\
    
    \midrule
    \rowcolor{gray!9} \multicolumn{7}{c}{\textit{Large Size Foundation Models ($>$ 100B)}} \\
    \midrule
    OpenAI-o3 & 70.5 & 58.1 & 49.7 & 20.2 & 66.7 & 71.7\\
    Claude-4-Sonnet & 68.3 & 29.1 & 12.2 & 20.3 & 64.6 & 61.7\\
    DeepSeek-V3.1 & 63.1 & 49.2 & 30.0 & 29.8 & 71.0 & 61.2 \\
    DeepSeek-V3.2 & 63.5 & 47.9 & 40.1 & 27.2 & 71.0 & - \\
    Kimi-K2 & 57.7 & 28.8 & 14.1 & 18.1 & 50.0 & 63.0 \\
    \midrule
    \rowcolor{gray!9} \multicolumn{7}{c}{\textit{Medium Size Deep Research Agents (30B-72B)}} \\
    \midrule
    MiroThinker-32B-DPO-v0.1 & 57.3 & 17.0 & 13.0 & 11.8 & - & 49.3 \\
    ASearcher-Web-QwQ& 52.8 & 15.6 & 5.2 & 12.5 & 42.1 & 34.3 \\ 
    WebDancer-QwQ & 51.5 & 18.0 & 3.8 & - & 39.0 & 47.9 \\
    WebSailor-32B  & 53.2 & 25.5 & 10.5 & - & 53.3 & -\\
    WebSailor-72B & 55.4 & 30.1 & 12.0 & - & 55.0 & -\\
    WebSailor-v2-30B-A3B (SFT) & 66.0 & 28.3 & 24.4 & 23.9 & - & -\\
    WebSailor-v2-30B-A3B (RL) & 74.1 & 44.1 & 35.3 & 30.6 & - & -\\
    WebShaper-32B & 52.4 & - & - & - & - & 51.4 \\
    WebShaper-72B & 60.0 & - & - & - & - & 52.2 \\
    Tongyi-DeepResearch & 70.9 & 46.7 & 43.4 & 32.9 & - & 72.2 \\
    \midrule
    \rowcolor{gray!9} \multicolumn{7}{c}{\textit{Small Size Deep Research Agents ($<$ 10B)}} \\
    \midrule
    MiroThinker-8B-DPO-v0.1 & 46.6 & 13.6 & 8.7 & - & - & 45.7 \\
    WebSailor-7B & 33.0 & 14.2 & 6.7 & - & 34.3 & - \\
    DeepDive-9B (SFT) & - & 15.7 & 5.6 & - & 35.0 & - \\
    DeepDive-9B (RL) & - & 15.1 & 6.3 & - & 38.0 & - \\
    WebExplorer-8B (RL)$^\dagger$ & \underline{47.6} & \textbf{31.8} & \textbf{14.6} & \underline{12.4} & \textbf{53.0} & 58.0 \\
    \rowcolor{blue!9}{\OurModel{}-8B (SFT)} & \underline{47.6} & 24.2  & 10.6 & 11.7  & 48.0 & \underline{60.0} \\
    \rowcolor{blue!9}{\OurModel{}-8B (DPO)} & \textbf{51.5} & \underline{26.6}  & \underline{12.8} & \textbf{13.8} & \underline{49.0} & \textbf{61.7} \\
    \bottomrule
    \end{tabular}
    }
    \end{spacing}
    \label{tab:main_results}
    \vspace{-0.15cm}
\end{table*}

\subsection{Experimental Setup}

We use Qwen3-8B as the base model and perform supervised fine-tuning on high-quality SFT data, followed by further optimization with constructed DPO preference pairs. Detailed training procedures and hyperparameter settings can be found in Appendix~\ref{appendix:model-training-details}.
During evaluation, we use the Serper API\footnote{\url{https://serper.dev}} for web search and Jina\footnote{\url{https://jina.ai}} for webpage content extraction. Retrieved webpages are summarized using DeepSeek-V3.1. 

We evaluate our method on six challenging benchmarks: GAIA~\citep{mialon2023gaia}, BrowseComp-zh/en~\citep{wei2025browsecomp, zhou2025browsecomp}, HLE~\citep{phan2025humanity}, XBench-DeepSearch~\citep{chen2025xbench}, and WebWalkerQA~\citep{wu2025webwalker}. As in prior deep research studies, we use the text subsets for GAIA and HLE. We compare our approach with foundation models and deep research agents.
Model performance is reported using the \textit{pass@1} metric.


\subsection{Main Results}
Table~\ref{tab:main_results} summarizes the main evaluation results of \OurModel{} across six challenging deep-research benchmarks.\vspace{-5pt}

\paragraph{Online RL is Not All You Need.}
\OurModel{} attains top-tier performance among agents of comparable scale and matches models with up to 30B parameters that are trained using online RL. Across all benchmarks, it delivers consistently strong results, indicating that advanced deep-research capabilities can be realized solely through offline training, and that online RL is not all you need for effective deep-research modeling.
On BrowseComp-zh, \OurModel{} achieves a score of 26.6, surpassing medium-scale systems such as WebSailor-32B (25.5) and approaching the performance of WebSailor-72B (30.1). On WebWalkerQA, it reaches 61.7, comparable to large foundation models including Claude-4-Sonnet (61.7) and DeepSeek-V3.1 (61.2). Importantly, these results are obtained without any online RL, whereas many competitors depend on extensive and costly online fine-tuning.
Among agents below 10B parameters, \OurModel{} consistently ranks at the top, significantly outperforming open-source baselines such as ASearcher and DeepDive on BrowseComp-zh, BrowseComp-en, and XBench-DS. All improvements over baseline models are statistically significant ($p<0.05$).\vspace{-5pt}


\paragraph{\OurModel{} Can Generalize Effectively.}
Although the training data primarily targets deep search tasks, \OurModel{} exhibits strong generalization across diverse task categories. For example, on the HLE benchmark, \OurModel{} (DPO) attains a score of 13.8, matching or surpassing models trained with online reinforcement learning, including WebExplorer-8B and ASearcher-Web-QwQ. This result is particularly significant because HLE involves reasoning patterns distinct from standard search tasks. We attribute this success to the model's ability to internalize high-level complicated problem-solving strategies, rather than merely memorizing search patterns. The consistent performance beyond the training domain underscores the effectiveness of our offline preference-based approach, showing that high-quality preference data can yield robust and transferable reasoning capabilities.

\paragraph{Offline RL Training Achieves Competitive Results.}
The comparison between the SFT and DPO variants highlights the benefits of our offline DPO methodology. DPO consistently enhances performance across all benchmarks: +3.9 on GAIA, +2.4 on BrowseComp-zh, +2.2 on BrowseComp-en, +2.1 on HLE, +1.0 on XBench-DeepSearch, and +1.7 on WebWalkerQA. 
These consistent gains confirm that advanced deep-research competency can be attained through offline preference optimization. 
Moreover, our offline framework incurs no API costs during training, offers greater stability, and provides a cost-efficient paradigm for academic researchers, advancing our goal of enabling accessible development of deep-research agents.

\section{Empirical Study}

This section presents an empirical analysis addressing central aspects of our approach: the quality of generated training data, the impact of context window size, and the effect of model scaling.

\subsection{Quality Analysis of Synthesized Data}

To rigorously evaluate the quality of data generated by \OurFramework{}, we constructed multiple SFT training sets with different sizes. These datasets were used to train the Qwen3-8B model, and model performance was evaluated on the BrowseComp-ZH benchmark. For comparison, we use SFT-only versions of WebExplorer-8B (trained on 13k samples) and DeepDive-9B (trained on 858 samples) as our baselines.

\begin{figure}[t]
    \centering
    \begin{minipage}{0.45\linewidth}
    \centering
    \includegraphics[width=0.9\linewidth]{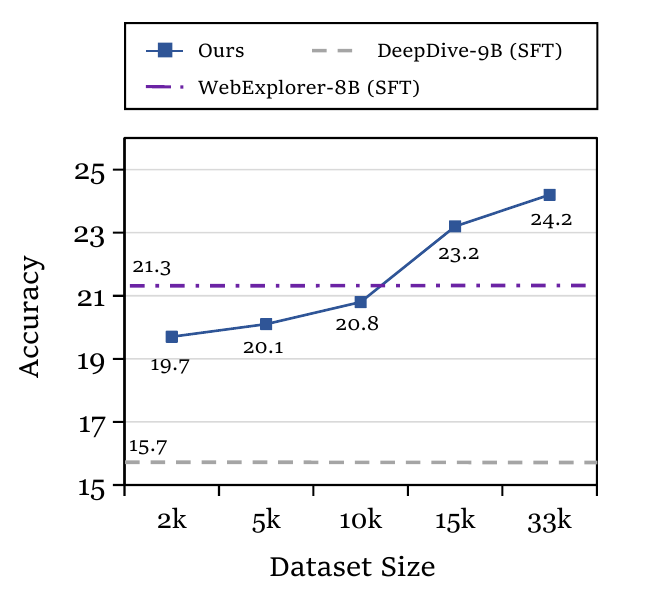}
    \vspace{-0.2cm}
    \caption{Pass@1 accuracy (\%) on BrowseComp-zh for different SFT models and dataset sizes. For our model, all tasks are generated by \OurFramework{}.}
    \label{fig:data-quality-analysis}
    \end{minipage}
    \hfill  
    \begin{minipage}{0.45\linewidth}
    \centering
    \includegraphics[width=0.9\linewidth]{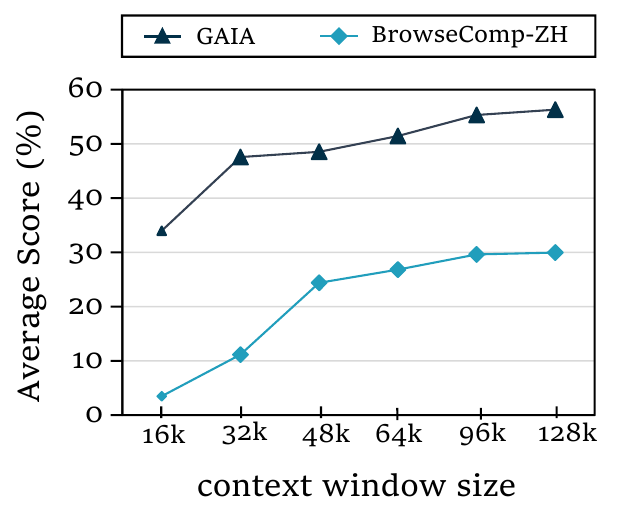}
    \caption{\OurModel{} test accuracy (\%) on GAIA and BrowseComp-ZH with different context window sizes.}
    \label{fig:token-context}
    \end{minipage}
    \hfill
\vspace{-0.2cm}
\end{figure}

Figure~\ref{fig:data-quality-analysis} shows that accuracy scales steadily with dataset size. Notably, with only 2k samples, our model achieves 19.7\% accuracy, substantially outperforming DeepDive-9B (15.7\% with 858 samples). This highlights the high quality and efficiency of data generated by \OurFramework{}.

\subsection{Effect of Different Context Window Sizes}

To assess how context length influences the performance of deep research models, we systematically evaluate \OurModel{} across GAIA and BrowseComp-ZH, under context windows from 16k to 128k tokens. 
As illustrated in Figure \ref{fig:token-context}, accuracy increases consistently with larger context windows. On GAIA, performance rises from 33.98\% at 16k to 56.31\% at 128k. The effect is even more pronounced on the more demanding BrowseComp-ZH benchmark, where accuracy improves from 3.48\% to 29.97\%. Notably, a 32k context window yields only about 35\% of the performance achieved with the 128k setting.
These findings underscore the critical role of extended context capacity in enabling complex, multi-hop reasoning and the integration of dispersed evidence, capabilities essential for advanced research-oriented tasks.

To study the effect of model scaling, we compare \OurModel{} of varying sizes (4B, 8B, and 14B parameters), trained on \OurDataset{} based on the Qwen3 series of models~\citep{yang2025qwen3}, and evaluate their performance on GAIA and BrowseComp-ZH.

\subsection{Effect of Model Scaling}
\begin{minipage}[t]{0.48\textwidth}
\textbf{Effect of Model Scaling.} 
As shown in Table~\ref{tab:model-scaling}, accuracy consistently improves as model size increases. The largest model, \texttt{\OurModel{}-14B} (SFT), achieves 51.4\% and 24.6\% on GAIA and BrowseComp-ZH, respectively. Applying DPO training yields additional gains across all sizes; \texttt{\OurModel{}-14B} with DPO reaches 54.4\% on GAIA and 27.0\% on BrowseComp-ZH, highlighting the effectiveness of offline DPO.
\end{minipage}%
\hfill
\begin{minipage}[t]{0.48\textwidth}
    \vspace{-2.2em}
    \centering
    \captionof{table}{Pass@1 accuracy (\%) of different Model sizes on GAIA and BC-zh. Larger models yield better accuracy on both benchmarks.}
    \scalebox{0.77}{
    \begin{tabular}{
        p{0.5\textwidth}
        p{0.25\textwidth}
        p{0.37\textwidth}
        }
    \toprule
    \textbf{Model Size} & \textbf{GAIA} & \textbf{BrowseComp-ZH} \\
    \midrule
    \texttt{\OurModel{}-4B} (SFT) & 37.8 & 18.3 \\
    +DPO & 43.7 \textcolor{seedblue}{+5.9} & 18.7 \textcolor{seedblue}{+0.4} \\
    \texttt{\OurModel{}-8B} (SFT)   & 47.6  & 24.2 \\
    +DPO & 51.5 \textcolor{seedblue}{+3.9} & 26.6 \textcolor{seedblue}{+2.4} \\
    \texttt{\OurModel{}-14B} (SFT)  & 51.4  & 24.6 \\
    +DPO & 54.4 \textcolor{seedblue}{+3.0} & 27.0 \textcolor{seedblue}{+2.4} \\
    \bottomrule
    \end{tabular}
    }
    \label{tab:model-scaling}
\end{minipage}

\section{Related Work}
\label{sec:related-work}

\subsection{Deep Research Agents}

Deep research agents have rapidly emerged as a new frontier in LLMs~\citep{openaideepresearch, team2025kimi, hu2025owl, fang2025cognitive, qiu2025alita, team2025mirothinker}, delivering strong performance on complex reasoning benchmarks such as GAIA~\citep{mialon2023gaia}. A growing body of work focuses on large-scale web-based exploration, exemplified by the Tongyi Deep Research series~\citep{wu2025webdancer, li2025websailor, li2025websailorv2}, WebExplorer~\citep{liu2025webexplorer}, and ASearcher~\citep{gao2025beyond}, which exhibit advanced browsing and information synthesis capabilities on benchmarks like BrowseComp~\citep{wei2025browsecomp}.

Most advanced systems heavily rely on online RL to achieve robust performance in deep research tasks. However, these methods are computationally expensive and resource-intensive. In contrast, offline learning remains underexplored, largely due to the limited availability of high-quality data. To address this gap, we introduce the first large-scale deep research SFT and DPO datasets designed to support efficient offline training for deep research agents.

\subsection{Deep Search Task Synthesis}

Deep search tasks differ fundamentally from standard web-based QA benchmarks such as HotpotQA~\citep{yang2018hotpotqa} and 2WikiMultihopQA~\citep{ho2020constructing}. Unlike conventional benchmarks, these tasks require reasoning under high uncertainty and sustained interaction with real-world search environments. However, high-quality data for deep search is currently scarce. Therefore, it is essential to automatically synthesize large-scale, diverse, and high-fidelity tasks to develop models with advanced web-based reasoning capabilities.
Several frameworks have been proposed to support such automatic synthesis. WebSailor~\citep{li2025websailor} and WebShaper~\citep{tao2025webshaper} focus on enhancing question diversity and uncertainty via fuzzing or set-theoretic composition. 
ASearcher~\citep{gao2025beyond} and DeepDive~\citep{lu2025deepdive} both use offline Wikipedia data to construct complex research tasks. They employ techniques like LLM agents and knowledge graph walks for complex query generation, and have publicly released their code and synthesized tasks.

Despite these advances, most existing frameworks rely on intricate preprocessing workflows and rarely release complete pipelines or datasets. Here, we present a simple and efficient framework that produces challenging deep search questions directly, without extensive preprocessing. All code and data are fully released to enable transparent and reproducible research.

\section{Conclusion}
This work addresses three critical challenges in deep research agent development: (1) the lack of efficient open-source data synthesis frameworks, (2) the shortage of high-quality trajectory data, and (3) the prohibitive cost of online reinforcement learning.
We show that online reinforcement learning is not essential for building strong deep research agents by establishing a fully offline and cost-effective training paradigm.
To alleviate data scarcity and training costs, we introduce a fully open-source resource suite: \OurFramework{}, a simple yet effective framework for synthesizing large-scale and diverse deep research tasks without heavy preprocessing; \OurDataset{}, which includes 66k QA pairs, 33k supervised fine-tuning trajectories, and 21k DPO preference pairs; and \OurModel{}, an 8B-parameter agent trained entirely offline.
Experiments across six benchmarks demonstrate that our offline approach matches or outperforms online RL-based systems, significantly lowering the barrier to developing high-quality deep research agents.

\section{Limitation}

Our offline approach achieves strong performance, but some limitations remain. Even with a 128k context window, certain complex queries (e.g., in BrowseComp-en) may exceed token limits, hindering full evidence integration; external memory or summarization may help. Additionally, our data primarily targets web search scenarios and does not yet cover domains like scientific literature or multimodal tasks, which we plan to explore in future work.

\clearpage

\bibliographystyle{plainnat}
\bibliography{main}

\newpage

\setcounter{section}{0}
\renewcommand{\thesection}{\Alph{section}}

\section{Appendix}
\label{sec:appendix}


\subsection{Details of Estimating API Cost}
\label{appendix:api-cost}
Following the official pricing\footnote[1]{https://serper.dev/billing}, the cost of 1K API calls is set to \$1. 
In Table~\ref{tab:opensource_comparison}, the total API cost is computed by assuming 15 search API calls per deep search task and using officially reported training steps and parameters, with the exception of ASearcher-v1. For ASearcher-v1, the average number of search queries during the initial training phase is set to 3.5. Some models lack reported cost values due to the absence of detailed training information, such as total steps or batch size. The values presented here estimate only the cost of training the primary model using online reinforcement learning methods and exclude any exploratory or analytical experiments. Consequently, the actual API costs are likely to be considerably higher than these estimates.

\subsection{Details of Task Synthesis Pipeline}
\label{appendix:data_generation_pipeline}
\subsubsection{Full Pipeline}
Algorithm~\ref{alg:deepforge_data_synthesis_pipeline} shows the pseudocode of the full pipeline of the task synthesis process. It consists of two main stages: (1) Scalable Entity Expansion and (2) Complex Question Generation.

\begin{spacing}{1.05}
\begin{algorithm*}[h]
  \caption{DeepForge Data Synthesis Pipeline}
  \KwIn{seed noun batch size $B$, max workers $W$, entity exploration depth distribution $P_\text{depth}$}
  \KwOut{High-difficulty multi-hop QA dataset $\mathcal{D}$}
  \SetKwFunction{GenNouns}{GenerateRandomNouns}
  \SetKwFunction{SearchURLs}{SearchURLs}
  \SetKwFunction{ExtractEntities}{ExtractEntitiesFromHTML}
  \SetKwFunction{Explore}{ExploreEntityGraph}
  \SetKwFunction{GenQA}{GenerateQAPair}
  \SetKwFunction{Prune}{PruneClues}
  \SetKwFunction{Validate}{ValidateQuality}
  \BlankLine
  \textbf{Stage 1: Scalable Entity Expansion}\\
  \For{batches until noun pool size target reached}{
      $\mathcal{N} \leftarrow$ \GenNouns{$B$} 
      \ForEach{$n \in \mathcal{N}$ \textbf{in parallel with} $W$ workers}{
          $\mathcal{U}_n \leftarrow$ \SearchURLs{$n$}
          \ForEach{$u \in \mathcal{U}_n$}{
              $c \leftarrow \text{html2text}(u)$
              $\mathcal{E}_u \leftarrow$ \ExtractEntities{$c$}
              Keep only long-tail entities; append to pool $\mathcal{E}$
          }
      }
  }
  Deduplicate and noise-filter $\mathcal{E}$ to form seed entity set $\mathcal{S}$
  \BlankLine
  \textbf{Stage 2: Complex Question Generation}\\
  \ForEach{seed entity $e \in \mathcal{S}$ \textbf{in parallel with} $W$ workers}{
      $d_e \sim P_\text{depth}$ \tcp*{Sample entity graph exploration depth from the user-provided distribution}
      $G_e \leftarrow$ \Explore{$e$, depth $= d_e$} \\
      $(q, a) \leftarrow$ \GenQA{$G_e$} \\
      $(q', a') \leftarrow$ \Prune{$q, a, G_e$} \\
      Append $(q', a')$ to set $\mathcal{D}$
  }  \tcp*{Prune removes overly specific clues while preserving solvability}
  \BlankLine
  Return final dataset $\mathcal{D}$ of hard, multi-hop, high-uncertainty QA pairs
  \label{alg:deepforge_data_synthesis_pipeline}
\end{algorithm*}
\end{spacing}

\subsubsection{Prompts for Each Stage}
Below is the prompt for generating initial noun batch:
\begin{PromptBox}{Prompt for GenerateRandomNouns}
Please generate \{batch\_size\} diverse Chinese or English nouns randomly, including abstract and concrete nouns.

They should cover various fields such as technology, geography, culture, art, nature, brands, movies, plants, animals, and organizations.

Requirements:
\begin{itemize}[leftmargin=16pt,itemsep=0pt,topsep=0pt]
    \item No duplicates allowed
    \item No numbering required
    \item Each noun should be on a separate line
    \item Only output the nouns themselves, without any explanations
    \item Do not include any nouns that have already appeared.
\end{itemize}
\end{PromptBox}

Below is the prompt for exploration step when generating entity graph:
\begin{PromptBox}{Prompt for in ExploreEntityGraph}
You are an agent that can search the web for information and crawl the webpage content of a url.

Your task is to gather ample information about the entity, including two core aspects:
\begin{enumerate}[leftmargin=16pt,itemsep=0pt,topsep=0pt]
\item The entity itself, such as its description, properties, relevant events, etc.
\item The relationships between the entity and other entities, such as its neighbors, etc.
\end{enumerate}

You can use the following tools to help you:
\begin{itemize}[leftmargin=16pt,itemsep=0pt,topsep=0pt]
    \item search\_google: to search the web for information
    \item crawl\_url\_content: to crawl the webpage content of a url
    \item search\_wiki: to search the wikipedia for information
\end{itemize}

Here are the tool schemas:

\texttt{\{tool\_schemas\}}\\

If you want to use tools, please output in json format (with name and arguments), enclosed by \texttt{<function\_call> </function\_call>} tags. 

Please output only one function call at a time.\\

You should first leverage tools to gather information about the entity, and finally output the final result in the following json format:
\begin{lstlisting}
<result>
{{
    "entity_self": [
        "value1",
        "value2",
        "value3"
    ],
    "entity_relations": {{
        "entity1": "relation1",
        "entity2": "relation2",
        "entity3": "relation3"
    }}
}}
</result>
\end{lstlisting}

For example, if the entity is "Nginx", the value in "entity\_self" can be
"Nginx is an HTTP web server, reverse proxy, content cache, and load balancer.",
"nginx-1.29.1 mainline has been released in 2025-08-13",
"nginx was publicly released in 2004"
"Nginx is free and open-source software, released under the terms of the 2-clause BSD license"

the value in "entity\_relations" can be "Igor Sysoev": "Nginx was created by Russian developer Igor Sysoev"
Now please start to gather information about the entity.

When generating properties and relations, do not use high-frequency entities instead of unpopular entities (such as low-profile players, niche research institutions, non-mainstream foundations, etc.)

Here is the entity you should explore: \{name\} \{description\}

\end{PromptBox}

Below is the prompt for generating QA pairs based on the entity graph:
\begin{PromptBox}{Prompt for GenerateQAPair}
  
    You are a helpful assistant that can generate a question-answer pair based on the given entity information.
    
    I have already collected a bunch of entities and their related information. For each entity, I have collected its properties and relations.

    The properties are used to describe the entity itself, and the relations are used to describe the relationships between the entity and other entities, which can be used to generate multi-hop search questions.

    Your task is to generate a challenging question-answer pair to test a model's ability to perform deep, multi-hop searches on the web. The question must force the model to navigate through information about obscure entities and cannot be answered using common knowledge alone.\\
    
    Core Principles:
    \begin{enumerate}[leftmargin=16pt,itemsep=0pt,topsep=0pt]
    \item[1.] Focus on Obscurity: The question must be centered around unpopular or lesser-known entities, rather than high-frequency entities.

    \item[2.] Promote Web Search: The question must be constructed so that answering it requires iterative web searches to verify relationships and properties. It should not be solvable through guesswork or general knowledge.
    \item[3.] Embrace Ambiguity \& Fuzziness: Descriptions must be vague and indirect. Avoid precise identifiers that act as direct lookup keys.
    \item[4.] Use: Ranges (e.g., "the 1970s," "a budget between \$10-20 million"), relative terms (e.g., "a short-lived show," "a moderately successful album"), and ambiguous descriptors (e.g., "a politician involved in an early environmental policy").
    \item[5.] Avoid: Exact dates (e.g. "2008 year"), specific numbers (e.g. "83rd minute"), well-known proper names (people, places, awards), clues that are easy to deduce, and unique superlatives (such as "the first," "the highest-grossing", "The city with the second highest wind speed"), which can be directly searched and found through search engines.\\
    \end{enumerate}

    Construction Guidelines:
    \begin{enumerate}[leftmargin=16pt,itemsep=0pt,topsep=0pt]
    \item[1.] Source Material: Use the provided information on an entity-including its properties and its relations to other entities-as the foundation for your question.
    \item[2.] Question Type: The final answer should be the name of the target entity.
    \item[3.] Language: The language of the generated question must match the language of the provided entity information, either Chinese or English.
    \item[4.] Describing the Entity: Build the question by weaving together vague descriptions of the entity's properties and its relations to other obscure entities. The path to the answer should require multiple logical "hops." For example:
    \end{enumerate}

    \begin{enumerate}[leftmargin=16pt,itemsep=0pt,topsep=0pt]
        \item[]    Hop 1: Identify Entity A based on its vague relation to a slightly more known concept.

    Hop 2: Discover that Entity A worked on a project with Entity B.

    Hop 3: Find a vague property of Entity B that leads to the final target, Entity C.\\
    \end{enumerate}
    Example of a Good vs. Bad Question:\\
    
    Bad (Too Direct): "What is the name of the player who died at the age of 44?" This uses a unique, precise fact that can be directly searched.

    Good (Vague \& Multi-Hop): "A supporting actor from a sci-fi film released in the late 80s later directed a made-for-TV movie that was nominated for a minor industry award in the mid-1990s. What is the name of this director?" This requires finding the actor, then their directing work, then filtering by a specific award timeframe.\\

    Your Output: Generate a single question-answer pair that adheres to all the principles above.
    
    Here are some examples for the ideal question-answer pair:

    \texttt{\{examples\}}\\
    
    Now please generate a question-answer pair based on the given entity graph.

    Here is the entity graph:

    \texttt{\{entity\_infos\}}\\
    
    Your output should be in the following format:
\begin{lstlisting}
<thinking>
[YOUR THINKING HERE, DESCRIBING WHY YOU WANT TO GENERATE THIS QUESTION-ANSWER PAIR]
</thinking>
<question>
[THE GENERATED QUESTION HERE]
</question>
<answer>
[THE GENERATED ANSWER HERE]
</answer>
\end{lstlisting}
\end{PromptBox}

\subsubsection{Question Example}
\label{appendix:deepforge_case_study}
Below is a question example synthesized by \OurFramework:

\begin{PromptBox}{Example}{
    \textbf{Question:}

A monster material, from the sequel to the acclaimed open-world game, is dropped by mummy-like enemies in the desert region. It's used to enhance clothing that glows in the dark and serves a core creative gameplay element. The official name of this element is used as the name of an exploration achievement in another turn-based RPG themed around interstellar travel. Unlocking this achievement requires crafting an item, including a consumable with a name derived from European legend. The recipe can be purchased in a spaceship setting with an Eastern sci-fi feel. What item does this consumable's recipe correspond to?

    \bigskip

    \textbf{Answer:}

Sword in the Stone

}
\end{PromptBox}

Here, the open-world game refers to ``The Legend of Zelda: Tears of the Kingdom'' and the turn-based RPG refers to ``Honkai: Star Rail''. The name of the exploration achievement is ``Surplus Construction'' and the consumable's name is derived from the European legend of the ``Sword in the Stone''. This task cannot be synthesized solely using offline Wikipedia page content, since Wikipedia entries for these games do not provide detailed information, such as the specific names and methods for obtaining certain in-game achievements. Access to richer, real-time Internet information is required to generate such tasks.


\subsection{Details of Agent Framework}
\label{appendix:agent-framework}

This section provides detailed specifications for each tool available to the agent. All tools follow a function-calling interface where the agent invokes tools using structured \texttt{<tool\_call>} tags and receives responses wrapped in \texttt{<tool\_response>} tags.

\paragraph{Search Tool.}
The \texttt{search} tool is used by the agent to acquire search 
engine results.

\textbf{Input:} A single search query string or a list of search query strings.

\textbf{Output:} Search results in JSON format, including page titles, snippets, and URLs.

\textbf{Usage:} The agent uses this tool to retrieve relevant webpages or information through web search.

\paragraph{Search Wiki Tool.}
The \texttt{search\_wiki} tool provides Wikipedia content for given entities.

\textbf{Input:} A list of entity names.

\textbf{Output:} JSON-formatted Wikipedia article contents, including titles and main text for each entity.

\textbf{Usage:} The agent uses this tool to obtain structured and authoritative background information about entities from Wikipedia.

\paragraph{Visit URLs Tool.}
The \texttt{visit\_urls} tool extracts specific information from webpages by summarizing their content according to a given query.

\textbf{Input:} A list of URLs and a query string specifying the target information.

\textbf{Output:} For each URL, a summary of information relevant to the query.

\textbf{Usage:} The agent uses this tool to focus on extracting query-relevant content from webpages.

\paragraph{Execute Code Tool.}
The \texttt{execute\_code} tool allows the agent to run Python code for computation.

\textbf{Input:} A string containing Python code.

\textbf{Output:} The output of code execution, including standard output, error messages, or return values.

\textbf{Usage:} The agent uses this tool for calculations, data processing, or other tasks requiring programmatic execution.

\subsection{SFT Data Filtering Pipeline}
\label{appendix:data-filtering}

To ensure high-quality training data, we adopt a robust filtering pipeline that systematically evaluates each trajectory across five quality criteria. These checks are applied sequentially, and a trajectory is discarded immediately upon failing any step:

\textbf{Invalid Tool Result Check:} This check identifies trajectories where the result of the current tool call is empty or contains an error message. Such issues are typically caused by external factors (e.g., network or server environment problems) rather than the agent's use of the tool itself.

\textbf{Format Error Check:} We verify that every assistant message conforms to the required output format (e.g., all reasoning steps are wrapped in \texttt{<think>} tags, tool calls are contained within \texttt{<tool\_call>} tags with proper JSON formatting).

\textbf{Token Length Check:} For each trajectory generated by \OurFramework, we ensure that the tokenized length is within the range of 8,192 to 131,072 tokens. Trajectories shorter than 8k tokens or longer than 128k tokens are filtered out. Short trajectories usually correspond to overly simple tasks that do not require deep research.

\textbf{Answer Correctness Check:} We utilize an LLM-based evaluator to determine whether the model's final answer is semantically equivalent to the ground truth. The evaluator is provided with the question, ground truth, and model response, and instructed to assess semantic equivalence while allowing flexibility in terms of formatting, capitalization, word order, and minor linguistic variations. Only trajectories with semantically correct answers are retained.

\textbf{Overall Quality Check:} Finally, we perform a comprehensive LLM-based quality assessment that examines each trajectory for: (1) tool call hallucinations (falsely claiming successful tool use); (2) mismatches between search keywords and returned results; (3) fabricated search results (claiming retrieval of information that was not actually obtained); (4) process fabrication (presenting plausible reasoning with incorrect underlying logic); (5) cognitive inconsistencies (persisting in incorrect assumptions in the face of contradictory evidence); and (6) logical errors within reasoning chains. The evaluator produces a structured judgment on whether the trajectory meets quality standards, and any trajectory failing this broad review is removed.

For the overall quality check stage, the prompt input to the LLM is as follows:

\begin{PromptBox}{Quality Check Prompt}
    
    You are a professional data quality evaluation expert. Please carefully 
    analyze the following SFT training data and determine whether its quality 
    is acceptable.\\

    \textbf{Data Content:}

    Question: \{question\}

    Ground Truth Answer: \{answer\}

    Model Response: \{model\_response\}\\
    
    \textbf{Conversation Trajectory:}

    \{trajectory\_text\}\\
    
    \textbf{Evaluation Criteria:}
    Please evaluate the data quality according to the following criteria. If \textbf{any} of the issues below are present, the data should be marked as ``Unqualified'':
    \begin{enumerate}[leftmargin=16pt,itemsep=0pt,topsep=0pt]
        \item \textbf{Tool Call Hallucination}: The model claims to have successfully invoked a tool, but the tool did not actually return any useful content.
        \item \textbf{Keyword-Result Mismatch}: The search keywords do not align with the actual content returned by the tool.
        \item \textbf{Fabricated Search Results}: The model claims to have retrieved search results when no such results were actually obtained.
        \item \textbf{Process Fabrication}: The model presents a seemingly correct solution process, but the underlying reasoning is incorrect.
        \item \textbf{Cognitive Inconsistency}: The model forms an incorrect initial assumption, encounters contradictory information during search, extracts the correct answer, but fails to update or revise its erroneous belief.
        \item \textbf{Logical Errors}: There are clear logical flaws in the reasoning process.\\
    \end{enumerate}

    \textbf{Evaluation Instructions:}
    \begin{itemize}[leftmargin=16pt,itemsep=0pt,topsep=0pt]
        \item Carefully analyze each evaluation criterion.
        \item If any quality issues are identified, clearly explain the specific problems.
        \item If the data quality is acceptable, explain why it passes the evaluation.
        \item If the ground truth answer is empty, you do \textbf{not} need to evaluate whether the model response matches the ground truth; only consider the other evaluation criteria.\\
    \end{itemize}
    
    \textbf{Output Format:}

    Please strictly follow the JSON format below:
    \begin{lstlisting}
    ```json
    {
        "quality_score": "Qualified" or "Unqualified",
        "reason": "A detailed explanation justifying why the data passes or fails the evaluation",
        "issues": ["Issue 1", "Issue 2", ...]  // If unqualified, list specific issues; if qualified, return an empty list
    }
    ```
    \end{lstlisting}

\end{PromptBox}

\subsection{Model Training Details}
\label{appendix:model-training-details}

For supervised fine-tuning, we set the maximum sequence length to 128k tokens to accommodate long-horizon reasoning and multi-step tool interactions. To enable efficient long-context training, we adopt \textsc{360-LLaMA-Factory}~\citep{zou360lf} with sequence parallelism. We use the \texttt{cosine\_with\_min\_lr} learning rate scheduler, with an initial learning rate of 3e-5, a minimum learning rate of 1e-7, and a warmup ratio of 0.1. The model is trained for 3 epochs, with global training batch size of 512.

For DPO training, we set the maximum sequence length to 48k tokens. We employ the same \texttt{cosine\_with\_min\_lr} scheduler, with a learning rate of 1e-6, a minimum learning rate of 1e-7, and a warmup ratio of 0.1. DPO training is performed for 1 epoch, with global training batch size of 512. We select the results after training 200 steps as the final DPO model.

\begin{figure*}[h]
    \centering
    \includegraphics[width=0.98\textwidth]{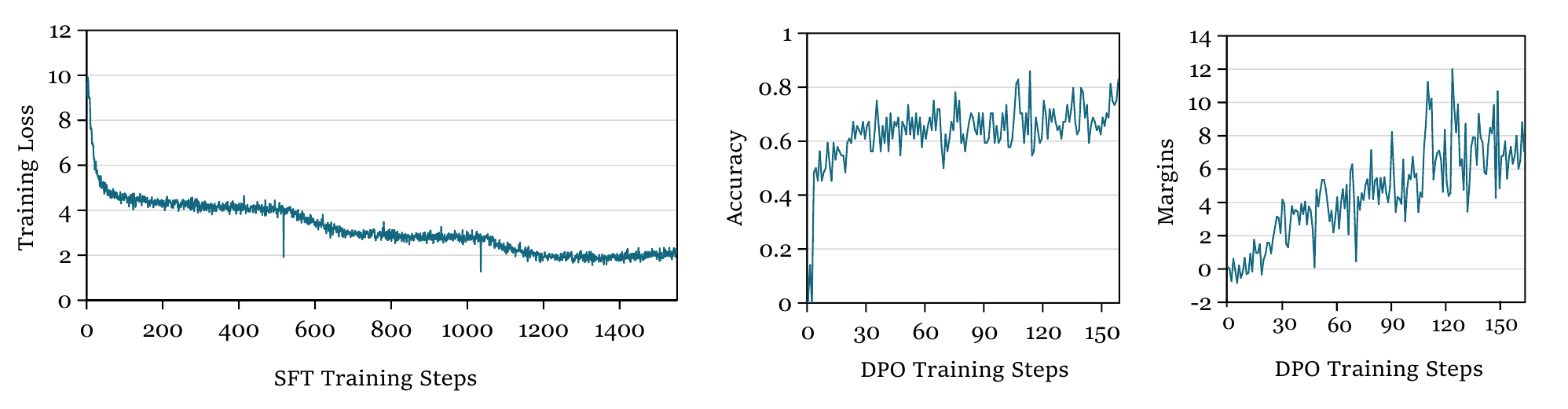}
    \caption{Training dynamics for supervised fine-tuning and DPO training. Left: SFT training loss curve. Middle: DPO training accuracy curve. Right: DPO training margin curve between positive and negative samples.}
    \label{fig:training_dynamics}
\end{figure*}

We determine all the hyperparameters by grid search. All training experiments are conducted on 64 NVIDIA H20 GPUs, each with 80\,GB of memory. Detailed training dynamics are shown in Figure~\ref{fig:training_dynamics}.

\subsection{Details about GRPO Training Efficiency}
\label{appendix:grpo-training-efficiency}
We set the train batch size to 128, each group consists of 8 rollouts. The experiments are conducted on 64 NVIDIA H20 GPUs. For each task, we set maximum processing time to 10 minutes. If timed out, we mark the task as failed.
For API cost estimation, we refer to the official pricing of Serper API. The cost of 1K API calls is set to \$1.

\subsection{Trajectory Example}
To illustrate how our agent framework works, we provide an example of a trajectory generated by our agent framework.
This example is sampled from GAIA benchmark, which demonstrates the agent's multi-step reasoning process and how it interacts with the environment with multiple tools to generate a final answer.

\begin{PromptBox}{Trajectory Example from GAIA}
\textbf{Question}

Use density measures from the chemistry materials licensed by Marisa Alviar-Agnew \& Henry Agnew under the CK-12 license in LibreText's Introductory Chemistry materials as compiled 08/21/2023.\\

I have a gallon of honey and a gallon of mayonnaise at 25C. I remove one cup of honey at a time from the gallon of honey. How many times will I need to remove a cup to have the honey weigh less than the mayonaise? Assume the containers themselves weigh the same.

\begin{center}
\begin{turnbox}
\centering \textbf{Turn 1}
\end{turnbox}
\end{center}

\textbf{Assistant Response:}

\texttt{<think>}

I need to solve this problem by comparing the weights of honey and mayonnaise at 25$\circ$C. The key is to determine their densities and calculate how many cups of honey need to be removed to make the remaining honey weigh less than the mayonnaise.\\

First, I need to find the densities of honey and mayonnaise at 25$\circ$C from the specified chemistry materials. I'll search for this information using the given context.

\texttt{</think>}

\texttt{<tool\_call>}
\begin{lstlisting}
{"name": "search", "arguments": {"query": ["density of honey at 25C", "density of mayonnaise at 25C", "honey mayonnaise density comparison", "CK-12 chemistry density measurements"]}}
\end{lstlisting}
    
\texttt{</tool\_call>}\\

    \textbf{Observation}
    \begin{lstlisting}
{"tool": "search", "result": [
    {"query": "CK-12 chemistry density measurements", "result": [{"link": "https://flexbooks.ck12.org/cbook/ck-12-chemistry-flexbook-2.0/section/3.11/primary/lesson/density-chem/", "snippet": "The relationship between the mass (m) of a material, its volume (V), and its density (D) is given by the formula D = m/V. This means that the density of a ...", "title": "Density - Definition, Formula, Units and Calculations | CK-12 ..."}, ...]},
    {"query": "honey mayonnaise density comparison", "result": [{"link": "https://www.quora.com/Which-has-the-higher-viscosity-mayonnaise-or-honey", "snippet": "The mayo has a higher viscosity. But when you drag a spoon through, the honey clearly offers more resistance. But the spoon will sink in the honey and not the ...", "title": "Which has the higher viscosity, mayonnaise or honey? - Quora"}, ...]},
     {"query": "density of mayonnaise at 25C", "result": [{"link": "https://kg-m3.com/material/mayonnaise-traditional", "position": 1, "snippet": "Mass density of Mayonnaise, traditional is 910 kg/m3. Mayonnaise, traditional specific gravity, mass- and volume calculator.", "title": "Mayonnaise, traditional density"}, ...]},
\end{lstlisting}

\begin{lstlisting}
    {"query": "density of honey at 25C", "result": [{"link": "https://kg-m3.com/material/honey", "position": 1, "snippet": "Honey density. Density of Honey is 1415 kg/m3 (between 1380 and 1450 kg/m$^{3}$ at 20 $\circ$C). *. Volumetric mass density of Honey (Food) in other popular units:.", "title": "Honey density"},
    ...]}]}
\end{lstlisting}
    \begin{center}
    \begin{turnbox}
    \centering \textbf{Turn 2}
    \end{turnbox}
    \end{center}

    \textbf{Assistant Response:}

\texttt{<think>}

I need to find the densities of honey and mayonnaise at 25$\circ$C. From the search results, I can see that:
\begin{itemize}[leftmargin=16pt,itemsep=0pt,topsep=0pt]
    \item Honey density at 25$\circ$C is approximately 1420 kg/m$^{3}$ or 1.42 g/cm$^{3}$
    \item Mayonnaise density at 25$\circ$C is approximately 910 kg/m$^{3}$ or 910 g/L
\end{itemize}

I should verify these values and make sure they're from the correct CK-12 materials. Let me search more specifically for the CK-12 density values.

\texttt{</think>}\\

\texttt{<tool\_call>}
\begin{lstlisting}
{"name": "search", "arguments": {"query": ["CK-12 Introductory Chemistry density honey mayonnaise", "LibreTexts CK-12 chemistry density measurements", "honey density 1.42 g/cm$^{3}$ CK-12", "mayonnaise density 910 kg/m$^{3}$ CK-12"]}}
\end{lstlisting}
\texttt{</tool\_call>}\\

    \textbf{Observation}
    \begin{lstlisting}
{"tool": "search", "result": [
\end{lstlisting}

\begin{lstlisting}
    {"query": "mayonnaise density 910 kg/m$^{3}$ CK-12", "result": [{"link": "https://kg-m3.com/material/mayonnaise-traditional", "position": 1, "snippet": "Mass density of Mayonnaise, traditional is 910 kg/m3. Mayonnaise, traditional specific gravity, mass- and volume calculator.", "title": "Mayonnaise, traditional density"}, ...]},
    {"query": "LibreTexts CK-12 chemistry density measurements", "result": [{"date": "Mar 20, 2025", "link": "https://chem.libretexts.org/Bookshelves/Introductory_Chemistry/Book%3A_Introductory_Chemistry_(CK-12)/03%3A_Measurements/3.11%3A_Density", "position": 1, "snippet": "Density is the ratio of the mass of an object to its volume. - Gases are less dense than either solids or liquids. - Both liquid and solid ...", "title": "3.11: Density - Chemistry LibreTexts"}, ...]},
    {"query": "CK-12 Introductory Chemistry density honey mayonnaise", "result": [{"date": "Jul 28, 2025", "link": "https://chem.libretexts.org/Bookshelves/Introductory_Chemistry/Introductory_Chemistry_(LibreTexts)/02%3A_Measurement_and_Problem_Solving/2.09%3A_Density", "position": 1, "snippet": "Density is a physical property found by dividing the mass of an object by its volume. Regardless of the sample size, density is always ...", "title": "2.9: Density - Chemistry LibreTexts"}, ...]},
    {"query": "honey density 1.42 g/cm$^{3}$ CK-12", "result": [{"date": "Sep 19, 2023", "link": "https://brainly.com/question/38015440", "position": 1, "snippet": "The volume of 12 grams of honey, given its density of 1.42 g/cm$^{3}$, is approximately 8.45 cm$^{3}$. We calculated this using the rearranged density ...", "title": "What is the volume of 12 grams of honey if its density is 1.42 g/cm$^{3}$?"},
    ...]}]}
\end{lstlisting}

    \begin{center}
    \begin{turnbox}
    \centering \textbf{Turn 3}
    \end{turnbox}
    \end{center}

    \textbf{Assistant Response:}

\texttt{<think>}

I need to find the density of honey and mayonnaise at 25$\circ$C from the CK-12 licensed materials in LibreText's Introductory Chemistry. The search results show some promising leads, particularly the "A Description of Matter - Introductory Chemistry" page which appears to be from a LibreTexts/CK-12 source and contains a table with the densities I need.\\

Let me visit that specific URL to get the exact density values from the licensed materials.

\texttt{</think>}\\

\texttt{<tool\_call>}
\begin{lstlisting}
{"name": "visit_urls", "arguments": {"urls": ["https://uen.pressbooks.pub/introductorychemistry/chapter/a-description-of-matter/"], "query": "density of honey and mayonnaise at 25$\circ$C in g/cm$^{3}$"}}
\end{lstlisting}
\texttt{</tool\_call>}

    \textbf{Observation}
    \begin{lstlisting}
{"tool": "visit_urls", "result": {"url": "https://uen.pressbooks.pub/introductorychemistry/chapter/a-description-of-matter/", "report": "Based on the webpage content, the density of honey and mayonnaise at 25$\circ$C is as follows: Honey: 1.420 g/cm$^{3}$; Mayonnaise: 0.910 g/cm$^{3}$. Summary: The information was found in Table 1.1 \"Densities of Common Substances\" within the webpage. The table explicitly lists the densities for both honey and mayonnaise at 25$\circ$C, measured in grams per cubic centimeter (g/cm$^{3}$)."}}
    \end{lstlisting}

    \begin{center}
    \begin{turnbox}
    \centering \textbf{Turn 4}
    \end{turnbox}
    \end{center}

    \textbf{Assistant Response:}
    
\texttt{<think>}

I have successfully found the densities of honey and mayonnaise from the specified LibreText's Introductory Chemistry materials. The density of honey is 1.420 g/cm$^{3}$ and the density of mayonnaise is 0.910 g/cm$^{3}$ at 25$^\circ$C.\\

Now I need to solve the problem:
\begin{itemize}[leftmargin=16pt,itemsep=0pt,topsep=0pt]
    \item I have a gallon of honey and a gallon of mayonnaise.
    \item I remove one cup of honey at a time.
    \item I need to find out how many times I need to remove a cup of honey for the remaining honey to weigh less than the mayonnaise.
    \item The containers themselves weigh the same, so I can ignore their weight in the comparison.
\end{itemize}

First, I need to convert gallons to cups to understand the volume relationship.
1 US gallon = 16 US cups.\\

The initial weight of the honey is: 1 gallon * density of honey = 16 cups * 1.420 g/cm$^{3}$
The initial weight of the mayonnaise is: 1 gallon * density of mayonnaise = 16 cups * 0.910 g/cm$^{3}$\\

Let x be the number of cups of honey removed.
The remaining volume of honey is: 16 cups - x cups.

The remaining weight of honey is: (16 - x) cups * 1.420 g/cm$^{3}$.\\

I need to find the smallest x such that the remaining honey weighs less than the mayonnaise.

So, I need to solve the inequality: (16 - x) * 1.420 $<$ 16 * 0.910\\

Let me calculate this step by step.

(16 - x) * 1.420 $<$ 16 * 0.910

16 * 1.420 - x * 1.420 $<$ 16 * 0.910

x * 1.420 $>$ 16 * 1.420 - 16 * 0.910

x * 1.420 $>$ 16 * (1.420 - 0.910)

x * 1.420 $>$ 16 * 0.510

x $>$ (16 * 0.510) / 1.420

x $>$ 8.16 / 1.420

x $>$ 5.747...\\

Since x must be a whole number (number of cups removed), I need to round up to the next whole number. So x = 6.\\

Let me verify this:

\begin{itemize}[leftmargin=16pt,itemsep=0pt,topsep=0pt]
    \item After removing 5 cups: (16 - 5) * 1.420 = 11 * 1.420 = 15.62 g
    \item After removing 6 cups: (16 - 6) * 1.420 = 10 * 1.420 = 14.20 g
    \item The mayonnaise weighs: 16 * 0.910 = 14.56 g\\
\end{itemize}

So, after removing 5 cups, the honey weighs 15.62 g, which is still more than the mayonnaise (14.56 g). After removing 6 cups, the honey weighs 14.20 g, which is less than the mayonnaise. Therefore, I need to remove 6 cups.\\

I will use the `execute\_code` tool to perform the calculation and verify.

\texttt{</think>}\\

\texttt{<tool\_call>}
\begin{lstlisting}
{"name": "execute_code", "arguments": {"code": "density_honey = 1.420\ndensity_mayo = 0.910\n\ngallons = 1\nvolume_in_cups = gallons * 16\n\ninitial_weight_honey = volume_in_cups * density_honey\ninitial_weight_mayo = volume_in_cups * density_mayo\n\nprint(f\"Initial weight of honey: {initial_weight_honey} g\")\nprint(f\"Initial weight of mayonnaise: {initial_weight_mayo} g\")\n\nx = (volume_in_cups * (density_honey - density_mayo)) / density_honey\n\nprint(f\"x = {x}\")\n\nimport math\nx_rounded = math.ceil(x)\nprint(f\"Number of cups to remove (rounded up): {x_rounded}\")\n\nremaining_weight_honey = (volume_in_cups - x_rounded) * density_honey\nprint(f\"Remaining weight of honey after removing {x_rounded} cups: {remaining_weight_honey} g\")\nprint(f\"Is remaining honey less than mayonnaise? {remaining_weight_honey < initial_weight_mayo}\")"}}
\end{lstlisting}
\texttt{</tool\_call>}\\``''

    \textbf{Observation}
    \begin{lstlisting}
{"tool": "execute_code", "result": "Initial weight of honey: 22.72 g Initial weight of mayonnaise: 14.56 g x = 5.746478873239436 Number of cups to remove (rounded up): 6 Remaining weight of honey 
\end{lstlisting}

\begin{lstlisting}
after removing 6 cups: 14.2 g Is remaining honey less than mayonnaise? True"}
\end{lstlisting}

    \begin{center}
    \begin{turnbox}
    \centering \textbf{Turn 5}
    \end{turnbox}
    \end{center}

    \textbf{Assistant Response:}

\texttt{<think>}
I have successfully found the densities of honey and mayonnaise from the specified LibreText's Introductory Chemistry materials.
- Density of honey at 25C: 1.420 g/cm$^{3}$
- Density of mayonnaise at 25C: 0.910 g/cm$^{3}$\\

I have also calculated the number of cups of honey to remove to make its weight less than the mayonnaise.

\begin{itemize}[leftmargin=16pt,itemsep=0pt,topsep=0pt]
    \item I need to remove 6 cups of honey.\\
\end{itemize}

The initial weight of honey is 22.72 g (1.420 g/cm$^{3}$ * 16 cups).

The initial weight of mayonnaise is 14.56 g (0.910 g/cm$^{3}$ * 16 cups).

After removing 6 cups of honey, the remaining weight is 14.2 g (10 cups * 1.420 g/cm$^{3}$).
14.2 g is less than 14.56 g, so the condition is met.\\

I have all the information needed to answer the question.

\texttt{</think>}\\

\texttt{<answer>}

You will need to remove 6 cups of honey to have the honey weigh less than the mayonnaise. Here is the calculation:

\begin{enumerate}[leftmargin=16pt,itemsep=0pt,topsep=0pt]
    \item \textbf{Densities:}
    \begin{itemize}[leftmargin=16pt,itemsep=0pt,topsep=0pt]
        \item Honey: 1.420 g/cm$^{3}$
        \item Mayonnaise: 0.910 g/cm$^{3}$
    \end{itemize}

    \item \textbf{Initial Volumes:}
\begin{itemize}[leftmargin=16pt,itemsep=0pt,topsep=0pt]
        \item 1 gallon of honey = 16 cups
        \item 1 gallon of mayonnaise = 16 cups
    \end{itemize}

    \item \textbf{Initial Weights:}
\begin{itemize}[leftmargin=16pt,itemsep=0pt,topsep=0pt]
        \item Weight of honey = 16 cups * 1.420 g/cm$^{3}$ = 22.72 g
        \item Weight of mayonnaise = 16 cups * 0.910 g/cm$^{3}$ = 14.56 g
    \end{itemize}

    \item \textbf{Let x be the number of cups of honey removed:}

\end{enumerate}

\begin{enumerate}[leftmargin=16pt,itemsep=0pt,topsep=0pt]
    \item[] \begin{itemize}[leftmargin=16pt,itemsep=0pt,topsep=0pt]
        \item The remaining volume of honey is (16 - x) cups.
        \item The remaining weight of honey is (16 - x) cups * 1.420 g/cm$^{3}$.
        \item We want the remaining weight of honey to be less than the weight of mayonnaise:

        $(16 - x) * 1.420 < 14.56$
    \end{itemize}

    \item \textbf{Solve for x:}
    \begin{itemize}[leftmargin=16pt,itemsep=0pt,topsep=0pt]
        \item $16 * 1.420 - x * 1.420 < 14.56$
        \item $22.72 - 1.420x < 14.56$
        \item $-1.420x < 14.56 - 22.72$
        \item $-1.420x < -8.16$
        \item $x > (-8.16) / (-1.420)$
        \item $x > 5.746$...
    \end{itemize}
    \item[6.] \textbf{Conclusion:}
    \begin{itemize}[leftmargin=16pt,itemsep=0pt,topsep=0pt]
        \item Since x must be a whole number (you can't remove a fraction of a cup), you need to remove 6 cups of honey.
    \end{itemize}
\end{enumerate}
\texttt{</answer>}

\end{PromptBox}






\end{document}